\documentclass[preprint,12pt]{elsarticle}

\usepackage{amssymb}
\usepackage{natbib}
\setcitestyle{numbers,square,comma}
\usepackage{makecell}

\begin{document}

\begin{frontmatter}



\title{GoogLe2Net: Going Transverse with Convolutions}


 \author[label1,label4]{Yuanpeng He}
 
 \affiliation[label1]{organization={Key Laboratory of High Confidence Software Technologies, Peking University},
	             city={Peking},
	             postcode={100871},
	             country={China}}
             
             

\affiliation[label4]{organization={School of Computer Science, Peking University},
	city={Peking},
	postcode={100871},
	country={China}}


\begin{abstract}
Capturing feature information effectively is of great importance in vision tasks. With the development of convolutional neural networks (CNNs), concepts like residual connection and multiple scales promote continual performance gains on diverse deep learning vision tasks. However, the existing methods do not organically combined advantages of these valid ideas. In this paper, we propose a novel CNN architecture called GoogLe2Net, it consists of residual feature-reutilization inceptions (ResFRI) or split residual feature-reutilization inceptions (Split-ResFRI) which create transverse passages between adjacent groups of convolutional layers to enable features flow to latter processing branches and possess residual connections to better process information. Our GoogLe2Net is able to reutilize information captured by foregoing groups of convolutional layers and express multi-scale features at a fine-grained level, which improves performances in image classification. And the inception we proposed could be embedded into inception-like networks directly without any migration costs. Moreover, in experiments based on popular vision datasets, such as CIFAR10 ($97.94$\%), CIFAR100 ($85.91$\%) and Tiny Imagenet ($70.54$\%), we obtain better results on image classification task compared with other modern models. 
\end{abstract}

%


\end{frontmatter}


\section{Introduction}
In recent years, we've witnessed a rapid advance of CNNs and this field is attracting more and more attention from researchers around the world. Noticeably, in order to meet demands of different vision tasks such as image classification, target tracking, image segmentation, skeleton extraction , facial recognition and image description, a large number of vision neural network models have been proposed \cite{DBLP:journals/tip/DongLDZ22, DBLP:conf/eccv/TianWKTI20, DBLP:conf/cvpr/Li0WH019, DBLP:conf/aaai/LuCYDC18, DBLP:conf/eccv/BennyW20, DBLP:journals/pami/ZhangLWWZ19}. And how to effectively extract information to satisfy demands of different kinds of vision tasks is still an open issue. Remarkably, capturing features from multiple scales to obtain more information has been a hot spot in computer vision-related fields which boosts performances of models.

The concept of multi-scale has already been introduced into deep learning-related fields \cite{DBLP:journals/mta/TangLX19, DBLP:conf/cvpr/AfifiDOB21, DBLP:journals/pami/BelongieMP02, DBLP:conf/cvpr/QiKG0WCLJ21} and its superiority was fully demonstrated by various applications. As a general rule, CNNs may acquire features utilizing convolutional kernels with different sizes from roughness to detail. Therefore, the key to boost performance of vision models is to devise a more efficient and effective policy of capturing features. And recently, on the basis of common residual block \cite{DBLP:conf/cvpr/HeZRS16}, a multi-scale architecture called Res2Net \cite{DBLP:journals/pami/GaoCZZYT21} is devised to better obtain and aggregate information at different scales. The idea of it resembles the one of Pyramid networks \cite{DBLP:conf/cvpr/LinDGHHB17} and the Res2Net block can continually enlarge the receptive filed through stacking $3$$\times$$3$ convolutional layers. Besides, the effectiveness of it is proved by the outstanding performance in diverse vision tasks.

\begin{figure}[htbp]
	\centering
	\includegraphics[width=0.6\linewidth]{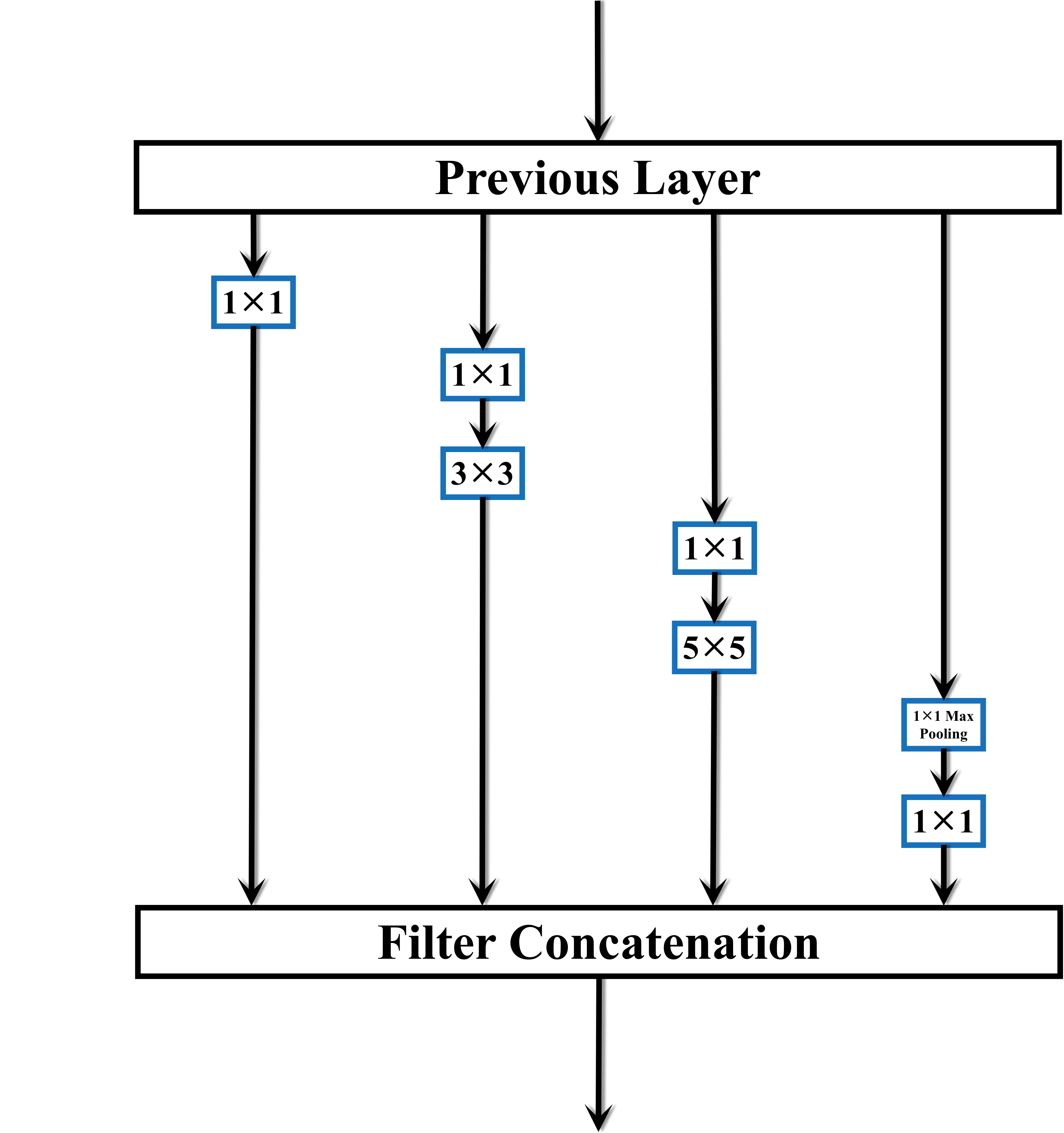}
	\caption{Original Inception from GoogLeNet}
	\label{Structure of Original Inception}
\end{figure}

Enlightened by the concept of Pyramid network and compositions of Res2Net block, we intend to generalize the idea of them to other networks which own relatively small parameter amount and similar architecture to ensure that the newly proposed block of network is efficient and modifications on it are straightforward. In order to fuse information more efficiently and acquire multi-scale features in larger receptive fields, we propose a novel GoogLe2Net based on GoogLeNet \cite{DBLP:conf/cvpr/SzegedyLJSRAEVR15}. The proposed GoogLe2Net has two versions which consists of residual feature-reutilization inception (ResFRI) and split-residual feature-reutilization inception (Split-ResFRI) respectively. About the model architecture, firstly, for the input layer, we adopt two disparate policies. For the first one, we utilize the original input layer from GoogLeNet without any changes; with respect to the second one, we split the input features into four different parts according to ratio of numbers of channels designed in GoogLeNet. The operation of split will significantly reduce the number of parameters and decrease training time a lot, however, which will also lead to a slight accuracy loss under some circumstances. For convolutional layers, different from existing inceptions with residual connections \cite{DBLP:conf/aaai/SzegedyIVA17}, we utilize the original structure of multi-scale of inceptions contained in GoogLeNet, which replaces the role of $3$$\times$$3$ convolutional layers in Res2Net to enhance the ability of network to extract more features from different scales. And the usage of $1$$\times$$1$ convolutional layers enables the model to capture stronger non-linearity in the same receptive field and avoids increasing calculation complexity too much. Therefore, we choose to remain consistent with GoogLeNet on the layout of convolutional layers. But for the improvement of performance, we construct transverse passages from the first to the last convolutional layer group, then information being processed can flow to next groups of convolutional layers. This operation enables information to be reutilized, in other words, the changes on the structure provide multi-scale feature extraction with a larger receptive field with respect to latter groups of convolutional layers, which makes up for the problem that the original structure does not utilize larger receptive field. Besides, in transverse passages, we adopt $1$$\times$$1$ convolutional layer to match features from channels between different groups of convolutional layers, which not only realizes the goal of construction of passages between groups of convolutional layers, but also reduces amount of parameters in comparison with $3$$\times$$3$ convolutional layer used in the structure of Res2Net. Besides, a residual connection is also added to the proposed inception to reduce difficulty of network optimization. Synthesizing the peculiarities mentioned before, the proposed network achieves relatively smaller model size and higher performance simultaneously. As a result, the ResFRI structure can be regarded as a satisfying solution in image classification task and innovation in CNN architecture.

\begin{figure*}[htbp]
	\begin{minipage}{1\linewidth}
		\centering
		\includegraphics[width=0.9\linewidth]{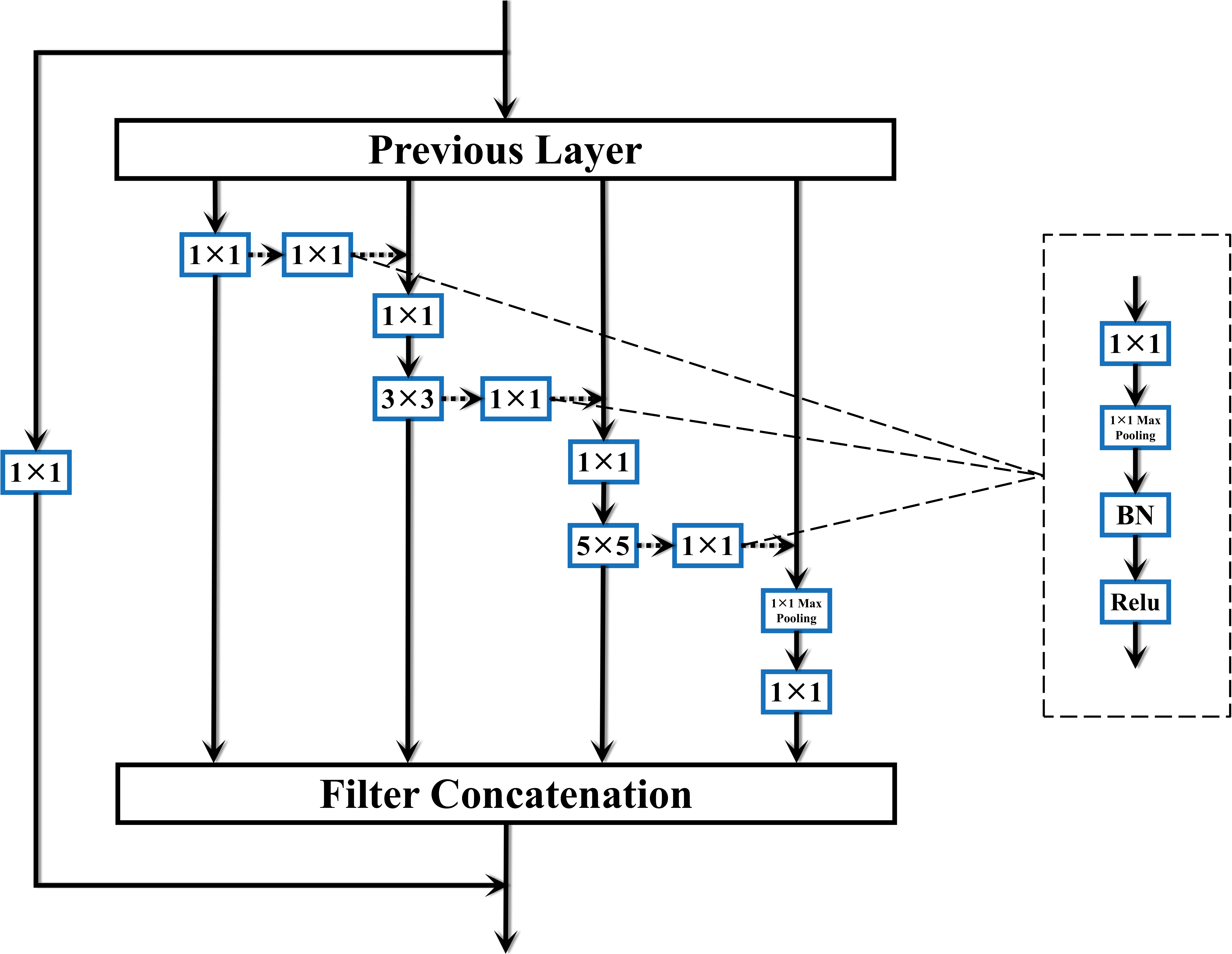}
		\caption{Residual Feature-Reutilization Inception from GoogLe2Net}
		\label{Structure of Residual Feature-Reutilization Inception}
	\end{minipage}
	\begin{minipage}{1\linewidth}
	\centering
	\includegraphics[width=0.9\linewidth]{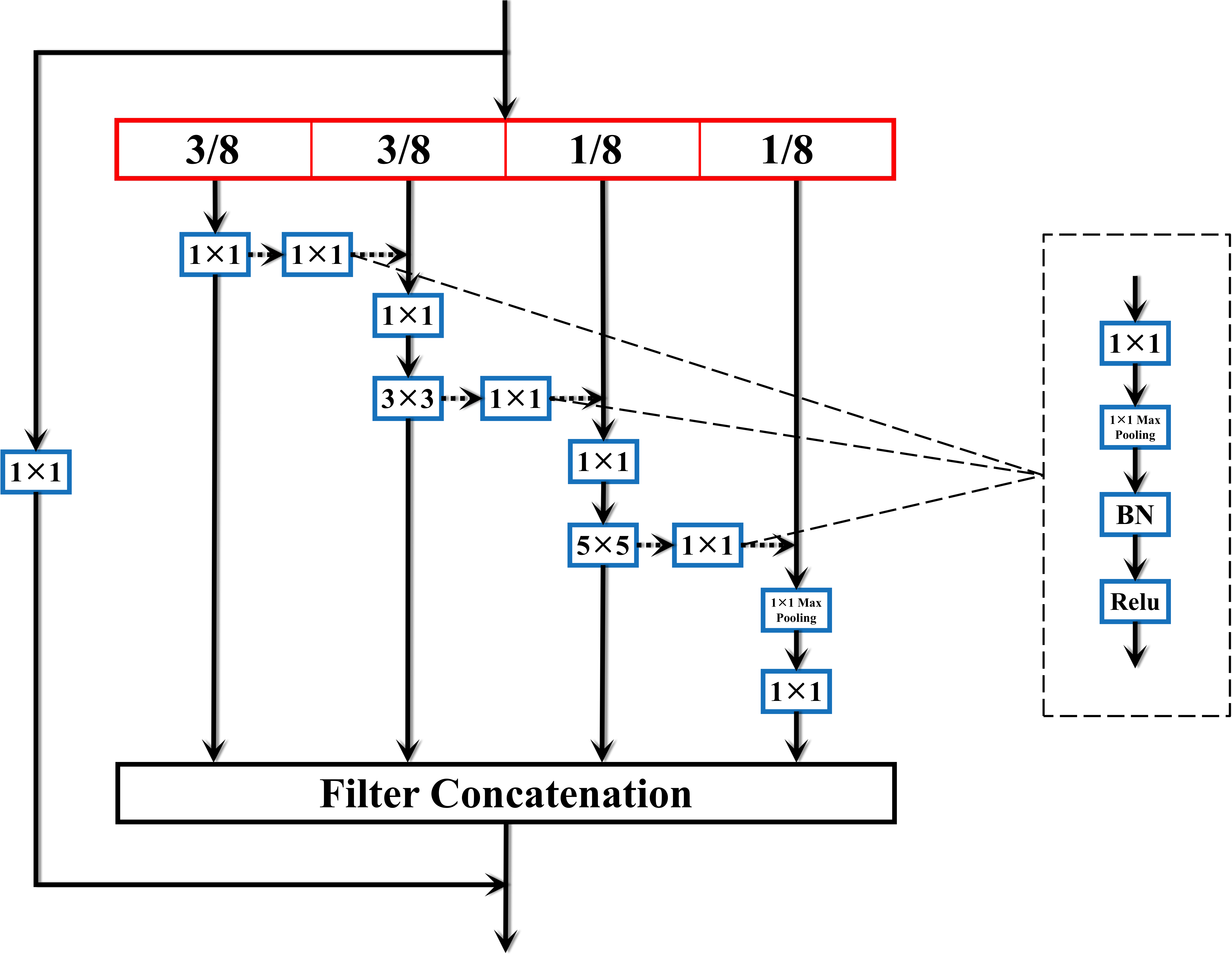}
	\caption{Split-Residual Feature-Reutilization Inception from GoogLe2Net}
	\label{Split structure of Residual Feature-Reutilization Inception}
\end{minipage}
\label{comparison}
\end{figure*}

All in all, GoogLe2Net combines features of multiple models and possess considerable advantages compared with other modern models. And the details of inception of GoogLeNet and ResFRI is provided in Fig.\ref{Structure of Original Inception} and Fig.\ref{Structure of Residual Feature-Reutilization Inception}, \ref{Split structure of Residual Feature-Reutilization Inception}. The main contribution of the ResFRI can be can be summed up in four points which are listed as below:
\begin{enumerate}
	\item GoogLe2Net explores influences brought by segmentation of information, which leads to reduction of parameter amount and training time. Besides, the loss of accuracy is also acceptable.
	\item The transverse passages in ResFRI enable the model to extract information in larger receptive fields and to fully utilize multi-scale features at fine-grained levels.
	\item The newly added residual connection in ResFRI could help GoogLe2Net optimize the whole network and gain better performance.
	\item GoogLe2Net investigates the effect of pruning and pruning ratio on the performance of this model, which inherits the idea provided by CondenseNet \cite{DBLP:conf/cvpr/HuangLMW18}.
\end{enumerate}

\section{Related work}
With the popularity of vision tasks, CNNs have made great progress \cite{DBLP:journals/tcsv/WenHL21, DBLP:journals/pami/HeGDG20, DBLP:journals/pami/LinGGHD20, DBLP:conf/miccai/RonnebergerFB15, DBLP:journals/pami/BadrinarayananK17, DBLP:journals/tip/LiuLLH22} and all of them contribute to the development of computer vision a lot. In order to improve performance of networks, researchers focus on adjusting depth and width of CNNs to better capture and process information. From the pioneering appearance of LeNet \cite{DBLP:conf/nips/CunBDHHHJ89} to some inspiring modern networks like AlexNet \cite{DBLP:conf/nips/KrizhevskySH12} and VGG \cite{DBLP:journals/corr/SimonyanZ14a}, both of them accelerate the advance of applications of neural networks. AlexNet \cite{DBLP:conf/nips/KrizhevskySH12} first adopts ReLu as activation function and utilizes dropout to ignore a part of neurons so that model overfitting can be avoided to some extent. Besides, AlexNet and its variant \cite{DBLP:conf/eccv/ZeilerF14} also achieve breakthroughs on network performance with respect to vision tasks, which is an outstanding progress compared with the methods proposed previously. And it's worth noting that there are lots of potentials on depth, width and receptive field of network which are also focuses in the future researches. In recent years, VGG-like networks \cite{DBLP:journals/corr/SimonyanZ14a, DBLP:conf/cvpr/Ding0MHD021} concentrate on stacking convolutional layers with small kernel size to enlarge size of receptive field and obtain information at a larger scale. And the work \cite{DBLP:conf/cvpr/Ding0MHD021} also introduces residual-like connections into framework of network to further enhance performance on vision tasks. More importantly, VGG outperforms AlexNet with less parameter amount, which mainly benefits from its ability to capture features at large scales. Compared with the proposed method in this paper, the receptive field of the two classical framework of networks are relatively fixed, which restricts their capability on processing information at diverse scales. Moreover, at that time, researchers also found that networks may encounter obstacles of overfitting, gradient vanishing and explosion while they're going deeper, which are difficulties need to be solved urgently.

Then, a classical neural network called GoogLeNet \cite{DBLP:conf/cvpr/SzegedyLJSRAEVR15} which was proposed by Christian Szegedy in $2014$. The module presented in Fig.\ref{Structure of Original Inception} is the basic structure of it. In order to avoid problem of overfitting and large calculation consumption, the inceptions contained in GoogLeNet improve performance of network and reduce parameter amount through combining convolutional layers on different magnitudes, which enhances its ability of more efficient utilization of computation resources and capture of more features at multi-scales. In the next year, another kind of network with residual connections called Resnet \cite{DBLP:conf/cvpr/HeZRS16} was proposed by Kaiming He to solve problem of network degradation and maintain accuracy when network becomes deeper. Following works like ResNext \cite{DBLP:conf/cvpr/XieGDTH17}, PreActResNet \cite{DBLP:conf/eccv/HeZRS16}, DenseNet \cite{DBLP:conf/cvpr/HuangLMW17} and Wide Residual Networks \cite{DBLP:conf/bmvc/ZagoruykoK16} prove the effectiveness and validity of residual connection, and as a result, the performances of networks are also guaranteed. With respect to vision task object detection, an efficient model called Pyramid networks \cite{DBLP:conf/cvpr/LinDGHHB17} was proposed and the concept of feature reutilization is introduced into modern neural network systems. And it can be roughly explained as that the high-level feature map will send the features back down and build the feature pyramid in reverse. Then the low-level feature map contains more location information, while the high-level feature map contains better classification information, combining the two level, the dual requirements of information for detection tasks can be satisfied. All in all, different models of networks contribute the development of CNNs through adjusting structures of them according to one or more specific properties of the networks.

\begin{figure}[h]
	\centering
	\includegraphics[width = 1\textwidth]{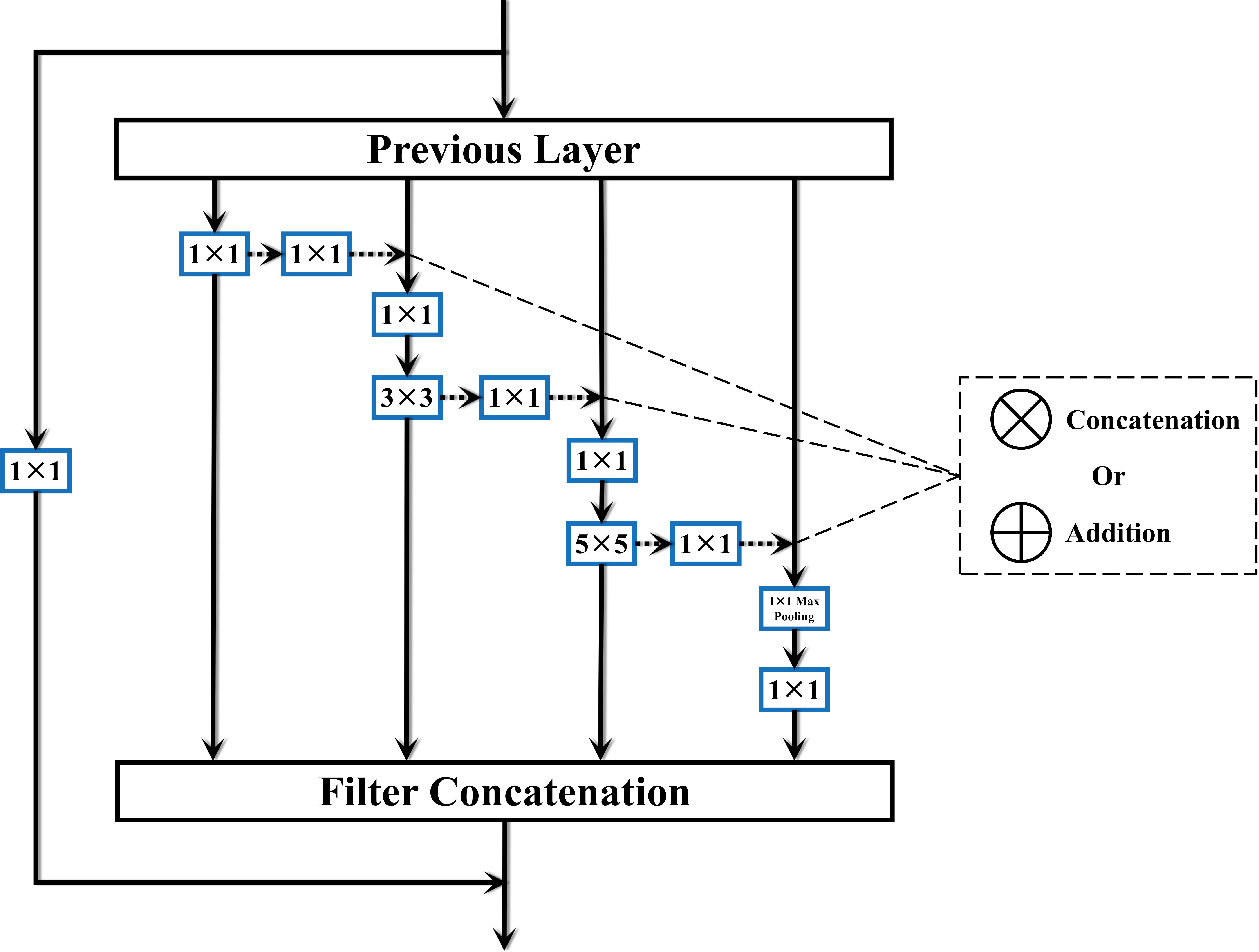}
	\caption{Details of transverse passages of GoogLe2Net.\\ \textbf{Note: The Split-ResFRI also adopts the same information interaction strategy as ResFRI}}
	\label{operation}
\end{figure}

\section{GoogLe2Net}
\subsection{Brief Introduction of Structure of GoogLe2Net}

The detail of ResFRI and Split-ResPRI are presented in Fig.\ref{Structure of Residual Feature-Reutilization Inception} and \ref{Split structure of Residual Feature-Reutilization Inception}. Suppose information from previous layer as $\xi_{Pre}$ and the operations of convolutional layers as $Conv$, the main difference of ResFRI (RI) and Split-ResFRI (SRI) in processing of information input can be defined as:

\begin{equation}
	\label{eq:alpha}
	\left\{
	\begin{array}{lcl}
		Conv(\xi_{Pre},\xi_{Pre},\xi_{Pre},\xi_{Pre}), &RI \\\\
		Conv(\gamma_{1},\gamma_{2},\gamma_{3},\gamma_{4})\\ \gamma_{1,2} = 3*\xi_{Pre}//8 \quad \gamma_{3,4} = \xi_{Pre}//8, &SRI
	\end{array}
	\right.
\end{equation}

Compared with the original structure of inception contained in GoogLeNet, residual connection and passages of information interaction between different groups of convolutional layers are added into ResFRI and Split-ResFRI. In order to reuse information, we construct transverse passages between adjacent groups of convolutional layers. Moreover, a residual connection is also devised to reduce difficulty of network optimization and to avoid problems like overfitting and abnormal gradients. Besides, to match feature channels between groups of convolutional layers and residual connection to final output, a structure consists of layers of $1$$\times$$1$ Convolutional layers, $3$$\times$$3$ MaxPool, BatchNorm and ReLu (cmbr) is utilized. It also further enhances extraction of information and realizes cross channel information combination and non-linear feature transference. And it's worth noting that the information combination is mainly achieved by adding or concatenating features and the operation is described in Fig.\ref{operation}. Suppose the information processed by former group of convolutional layer as $\delta$ and the information input to this group as $\kappa$, then the fusion of information between groups of convolutional layer can be defined as:
\begin{equation}
		\label{eq:alpha}
		\mathbb{F} =\left\{
		\begin{array}{lcl}
				Addition(cmbr(\delta), \kappa) \\
				Concat(cmbr(\delta), \kappa)
			\end{array}
		\right.
\end{equation}

Moreover, the comparison of performance and resource consumption between these methods can be found in the ablation study based on ResFRI.

To reduce consumption of computation resources, we discard the $3$$\times$$3$ convolutional layers designed by Res2Net and comply with the original design of inception of GoogLeNet. And we notice that the idea of connections between different groups of convolutional layers is very similar to the one of DenseNet \cite{DBLP:conf/cvpr/HuangLMW17}, the extra passages may help improve performance of network. However, \cite{DBLP:conf/cvpr/HuangLMW18} points out that the dense connections are actually redundant under certain circumstances and this phenomenon may reduce accuracy and efficiency of network. As a result, we prune newly-added passages of information transference except the residual connection in ResFRI to avoid unnecessary calculations and obtain higher accuracy. More specifically, we adopt unstructured pruning which trims the single weight and does not require a whole row of pruning. The advantage is that the original accuracy can be maintained, because structured pruning is easy to cut out those important weights. The tools of pruning is provided by Pytorch and unstructured pruning will abandon a part of weight parameters using mask matrices without changing the original size of models. For the filter concatenation ($\mathbb{FC}$) and synthesizing the operations defined above, suppose $Conv$ consists of $[\mathbb{C}_1,\mathbb{C}_2,\mathbb{C}_3,\mathbb{C}_4]$, it can be defined as:
\begin{equation}
		\label{eq:alpha}
		\mathbb{FC} =\left\{
		\begin{array}{lcl}
				Concat(\mathbb{C}_1(\xi_{Pre}),\mathbb{C}_2(\mathbb{F}(\mathbb{C}_1(\xi_{Pre})),\xi_{Pre}),\\ \mathbb{C}_3(\mathbb{F}(\mathbb{C}_2(\mathbb{F}(\mathbb{C}_1(\xi_{Pre})),\xi_{Pre})),\xi_{Pre}),\\ \mathbb{C}_4(\mathbb{F}(\mathbb{C}_3(\mathbb{F}(\mathbb{C}_2(\mathbb{F}(\mathbb{C}_1(\xi_{Pre})),\xi_{Pre})),\xi_{Pre})),\xi_{Pre})), \ RI\\\\
				Concat(\mathbb{C}_1(\gamma_1),\mathbb{C}_2(\mathbb{F}(\mathbb{C}_1(\gamma_1)),\gamma_2),\\ \mathbb{C}_3(\mathbb{F}(\mathbb{C}_2(\mathbb{F}(\mathbb{C}_1(\gamma_{1})),\gamma_{2})),\gamma_{3}),\\ \mathbb{C}_4(\mathbb{F}(\mathbb{C}_3(\mathbb{F}(\mathbb{C}_2(\mathbb{F}(\mathbb{C}_1(\gamma_{1})),\gamma_{2})),\gamma_{3})),\gamma_{4})), \ SRI
			\end{array}
		\right.
\end{equation}

The results of experiments in the following will prove the validity of pruning on diverse vision datasets.

\subsection{Other Important Settings of GoogLe2Net}
To ensure fair comparisons, the rest of settings of the whole network generally follow the principle formulated in GoogLeNet. And during the process of experiment, we notice that the MaxPool layers may hamper the network to capture information effectively and weaken performance of it, we argue that the MaxPool layers may destruct information contained in the low-resolution pictures instead of being helpful in extraction of features. Verified by experiments, we change the MaxPool layer into AvgPool layer eventually.

Argued by \cite{DBLP:conf/cvpr/HuangLMW18}, the dense connections may have negative impact on the process of learning and decrease accuracy of models. Therefore, we try to cancel some transverse passages to avoid too dense connections between adjacent groups of convolutional layers contained in the two version of GoogLe2Net utilizing different pruning ratio. Eventually we set the drop rate of passages of information transference to $0.7$ and $0$ on addition and concatenation version of ResFRI respectively, which can be defined as:
\begin{equation}
	\label{eq:alpha}
		Pruning\ Ratio =\left\{
		\begin{array}{lcl}
			 0.7,\quad   &Addition,RI \\
			 0,\quad   &Concatenation,RI
		\end{array}
		\right.
\end{equation}

With respect to Split-ResFRI, because of underlying performance loss which may be brought by segmentation of information, we set the pruning rate uniformly to $0$ in order to strengthen information interaction among groups of convolutional layers. And it is worth noting that when the classes contained in datasets are becoming more, we are supposed to reduce the amount of pruning to better promote information transference for the version of addition of ResFRI, which can be illustrated in the following experiments on vision datasets. In the last, the results in the part of ablation study will prove the effectiveness of these modifications based on ResFRI.

\section{EXPERIMENTS}
\begin{figure*}[htbp]
	\centering
	\includegraphics[width = 1\textwidth]{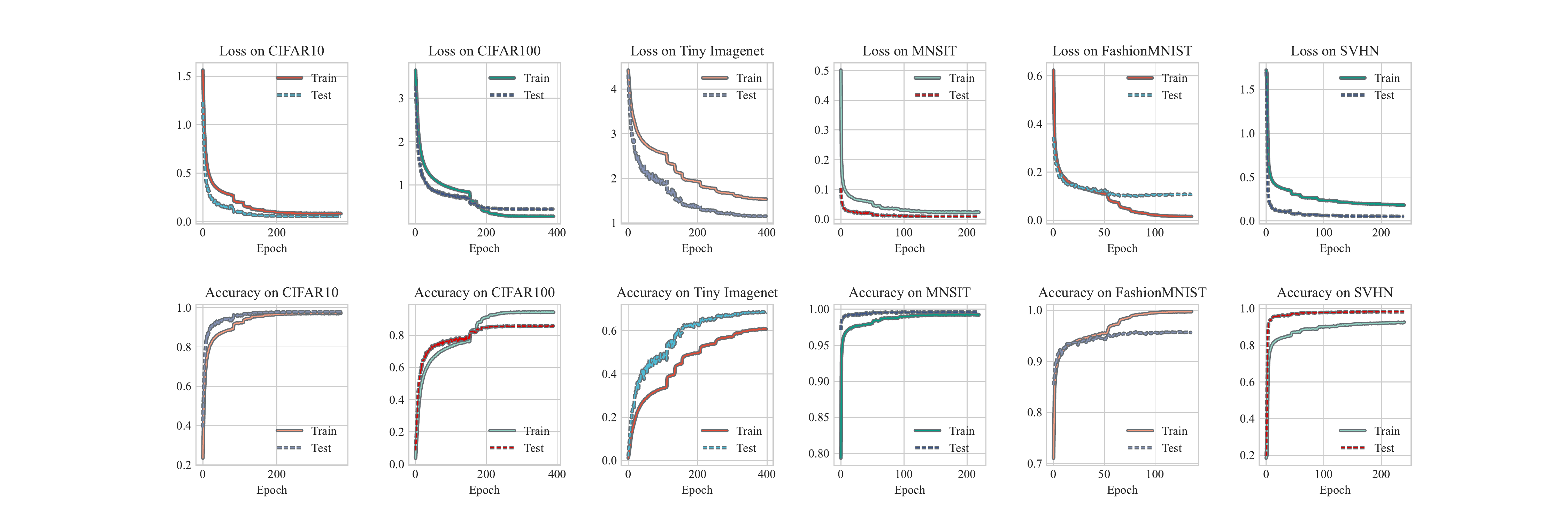}
	\caption{Loss and Accuracy of ResFRI-addition on Datasets}
	\label{tendency}
\end{figure*}
\subsection{Implementation Details}

We implement the whole framework of GoogLe2Net utilizing code framework provided by Pytorch. And in order to ensure fairness of comparison among different methods, we select experiment results of classical and newly proposed models without pre-training. Due to our limited computation resources, apart from necessary ablation experiments, we choose the task of image classification on the common datasets, such as CIFAR10, CIFAR100, Tiny Imagenet, MNIST, FashionMNIST and SVHN. Besides, in the process of training on one RTX $3060$ GPU, we use the optimizer SGD with momentum $0.9$, weight decay $0.0005$, batch size $64$ and data augmentation tools provided in packages of torchvision. Moreover, the initial learning rate is set to $0.01$ and it is reduced by half if validation loss does not decrease within $10$ epochs. And tendency of accuracy and loss in the training process of ResFRI is given in Fig.\ref{tendency}.

\subsection{Experiments on CIFAR-10}
\begin{figure*}[htbp]
	\centering
	\includegraphics[width = 1\textwidth]{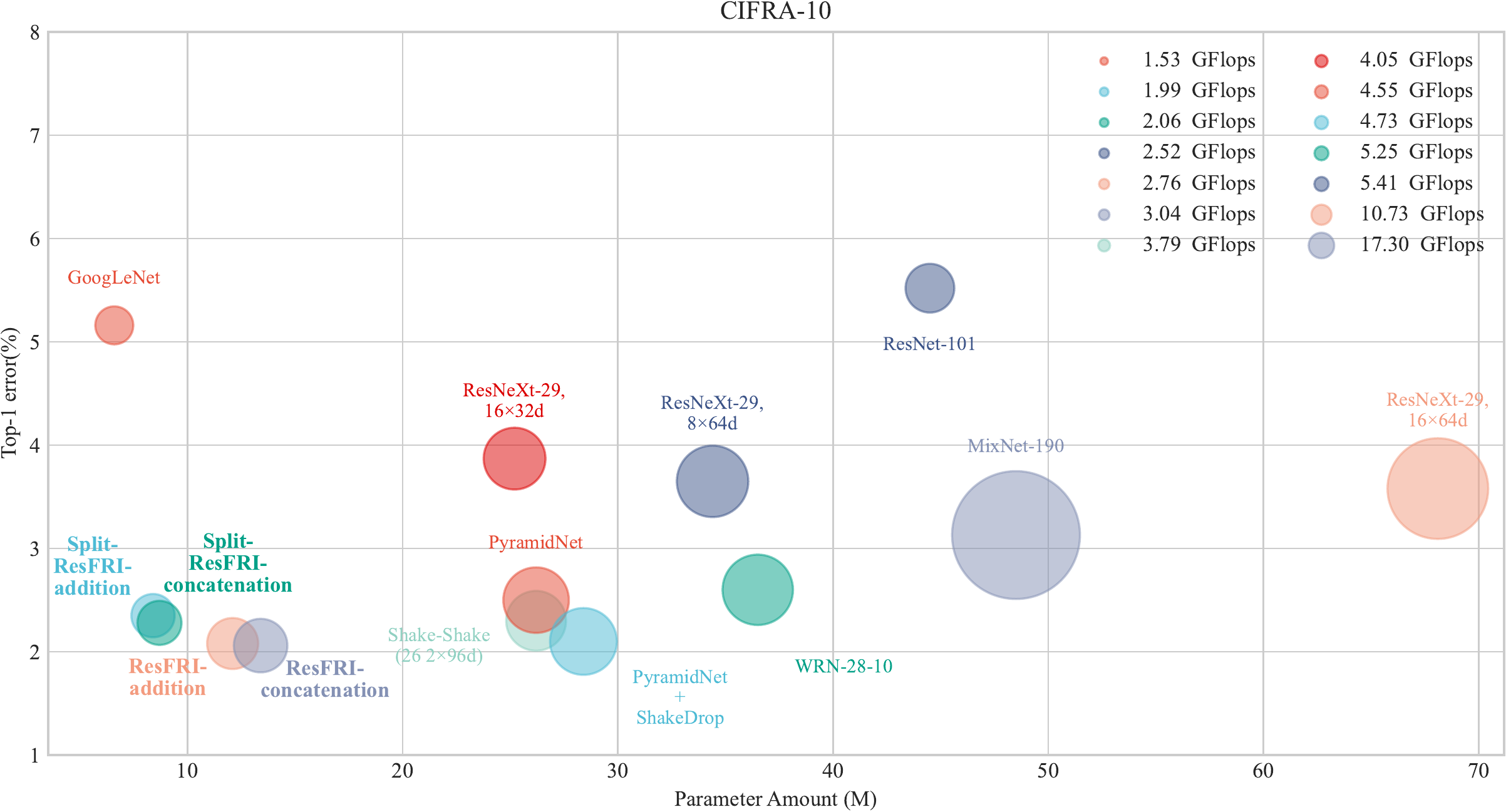}
	\caption{Comparisons of models on CIFAR10 Dataset }
	\label{CIFAR10_pop}
\end{figure*}

The CIFAR10 dataset contains $50$k training images and $10$k testing images from $10$ classes whose resolution is $32$$\times$$32$. And the detail results of comparisons of different models will be clearly provided in Table \ref{table1} and Fig.\ref{CIFAR10_pop}.

\begin{table*}[htbp]\footnotesize
	\centering
	\caption{Error rate (\%) and Model Size on the CIFAR-10 Dataset}
	\renewcommand\arraystretch{1.4} 
	\begin{tabular}{ccccccc} 
		\hline 
		\multicolumn{4}{l}{Model} & Flops&  Params  &    top-1 err. \\ \cline{1-7}
		\multicolumn{4}{l}{ResNet-101 \cite{DBLP:conf/cvpr/HeZRS16}}&2.52 GFlops &     44.5M    &      5.52        \\
		\multicolumn{4}{l}{GoogLeNet \cite{DBLP:conf/cvpr/SzegedyLJSRAEVR15}}&1.53 GFlops &     6.6M    &      5.16        \\
		\multicolumn{4}{l}{ResNeXt-29, 16$\times$32d \cite{DBLP:conf/cvpr/XieGDTH17}}&4.05 GFlops&     25.2M    &      3.87       \\
		\multicolumn{4}{l}{ResNeXt-29, 8$\times$64d \cite{DBLP:conf/cvpr/XieGDTH17}}&5.41 GFlops&     34.4M    &      3.65        \\
		\multicolumn{4}{l}{ResNeXt-29, 16$\times$64d \cite{DBLP:conf/cvpr/XieGDTH17}}&10.73 GFlops &     68.1M    &      3.58       \\
		\multicolumn{4}{l}{	CapsNet \cite{DBLP:conf/nips/SabourFH17}}&- &     -   &      10.6    \\
		\multicolumn{4}{l}{	DropConnect \cite{DBLP:conf/icml/WanZZLF13}}& -&     -   &      9.32      \\
		\multicolumn{4}{l}{	NIN + Dropout + Data Augmentation \cite{DBLP:journals/corr/LinCY13}}&- &     0.96M   &      8.81     \\
		\multicolumn{4}{l}{	RMDL \cite{DBLP:conf/icisdm/KowsariHBMB18}}&- &     -   &      8.74      \\
		\multicolumn{4}{l}{	FractalNet \cite{DBLP:conf/iclr/LarssonMS17}}&- &     38.6M   &      7.27      \\
		\multicolumn{4}{l}{	FitNet-LSUV \cite{DBLP:journals/corr/MishkinM15}}&- &     0.3M   &      6.06      \\
		\multicolumn{4}{l}{	SOPCNN \cite{DBLP:conf/mldm/Assiri19}}&- &     4.2MB   &      5.71      \\
		\multicolumn{4}{l}{DenseNet-BC (k=24) \cite{DBLP:conf/cvpr/HuangLMW17}}&- &    15.3M     &      5.19        \\
		\multicolumn{4}{l}{DPN-28-10 \cite{DBLP:conf/aaai/YangAZHZXLX20}}&- &     47.8M    &      3.65        \\
		\multicolumn{4}{l}{NASNet-A \cite{DBLP:conf/aaai/YangAZHZXLX20}}&- &      3.3M    &       3.41       \\
		\multicolumn{4}{l}{AmoebaNet-A \cite{DBLP:conf/aaai/YangAZHZXLX20}}&- &      4.6M    &       3.34       \\
		\multicolumn{4}{l}{AOGNet \cite{DBLP:conf/aaai/YangAZHZXLX20}}&- &     24.8M    &      3.27      \\
		\multicolumn{4}{l}{MixNet-190 \cite{DBLP:conf/aaai/YangAZHZXLX20}}&17.3 GFlops &     48.5M    &      3.13      \\
		\multicolumn{4}{l}{AmoebaNet-B \cite{DBLP:conf/aaai/YangAZHZXLX20}}&- &      34.9M    &       2.98       \\
		\multicolumn{4}{l}{OR-WideResNet \cite{DBLP:conf/cvpr/ZhouYQJ17}}&- &     18.2M    &       2.98       \\
		\multicolumn{4}{l}{WRN-28-10 \cite{DBLP:conf/icml/KwonKPC21}}&5.25 GFlops &    36.5M    &       2.6      \\
		\multicolumn{4}{l}{PyramidNet \cite{DBLP:conf/cvpr/LinDGHHB17}}&4.55 GFlops &    26.2M    &       2.5      \\
		\multicolumn{4}{l}{Shake-Shake (26 2x96d) \cite{DBLP:conf/iclr/ForetKMN21}}&3.79 GFlops &    26.2M    &       2.3      \\
		\multicolumn{4}{l}{PyramidNet+ShakeDrop \cite{DBLP:conf/iclr/ForetKMN21}}&4.73 GFlops &     28.4M   &      2.1       \\\cline{1-7}
		\multicolumn{4}{l}{ResFRI-addition}& 2.76 GFlops &      12.1M   &      2.08        \\
		\multicolumn{4}{l}{Split-ResFRI-addition}& 1.99 GFlops &      8.4M   &      2.35        \\
		\multicolumn{4}{l}{ResFRI-concatenation}&3.04 GFlops &     13.4M     &  \textbf{2.06} \\
		\multicolumn{4}{l}{Split-ResFRI-concatenation}&2.06 GFlops &     8.7M     &   2.28\\
		\hline
	\end{tabular}
	\label{table1}    
\end{table*}

It can be obtained that the ResFRI and Split-ResFRI achieve relatively satisfying performance on image classification task on CIFAR-10 dataset. Compared with traditional models like ResNet-101 and ResNeXt-29, ResFRI and Split-ResFRI have much better performance with much lower parameter amount. Although ResFRI-addition has $0.24$ GFlops and ResFRI-concatenation has $0.52$ GFlops higher than ResNet-101, we have a remarkable $3.44$\% and $3.46$\% performance gain on top-1 err while parameter amounts reduce by $32.4$M and $31.1$M. For Split-ResFRI, the version of addition has $0.53$ GFlops and $36.1$M parameters lower than ResNet-101, but we get $3.17$\% performance improvement.  Besides, Split-ResFRI-concatenation has $0.46$ GFlops  and $35.7$M parameters lower than ResNet-101. Both of the Split-ResFRIs have lower Flops and parameter amounts and achieve better results than ResNet-101. Compared with two versions of ResFRI, Split-ResFRIs sacrifice a little bit of precision in exchange for a considerable reduction in Flops and parameter amount. For ResNeXt-29, it outperforms ResNet-101 using larger model scales, but it still trails by at least $1.23$\% in comparison with ResFRI and Split-ResFRI. And with respect to GoogLeNet, no matter it is ResFRI or Split-ResFRI, we all have achieved performance leadership. It is worth noting that both versions of Split-ResFRIs have similar flops and parameter amounts to GoogLeNet, but still achieve a performance lead of over $2.8$ percentage. And with respect to CapsNet, DropConnect, NIN and RMDL, the four models reach a fairly satisfying level on small-size datasets like MNIST utilizing very small model scales, which partly outperforms many classical and novel methods including GoogLe2Net. However, all of the four models are not as good a performance as before in the more popular vision dataset, CIFAR-10, other modern models have overwhelming advantages compared with their results. Especially, the series of models belonging to ResFRI achieve at least a $6.39$\% performance lead.

Moreover, when encountering some newly proposed models, ResFRI and Split-ResFRI still prove their superiority on classification task. For DenseNet, it has a similar model scale to ResFRI-concatenation, but it has a $3.13$\% performance disadvantage in the final result. Besides, with respect to OR-WideResNet, it achieves a relatively satisfying results with acceptable model size. Compared with ResFRI and Split-ResFRI, its disadvantage is still significant with performance trailing by at least $0.63$\%. Then, WRN-28-10, PyramidNet and Shake-Shake($26$ $2$$\times$$96d$) all of them have higher flops and parameter amount than ResFRI and Split-ResFRI, but all of them achieve better accuracy except for Split-ResFRI-addition meanwhile. However, we want to point out that Split-ResFRI-addition has far less GFlops and parameter amount than the above model for comparison. Moreover, we notice that PyramidNet+ShakeDrop has a a very approximate performance ($-0.04$\%) to ResFRI-concatenation, which is a is a very competitive opponent. However, the cost of the combination of PyramidNet and ShakeDrop is $71.3$\% higher flops and $134.7$\% larger parameter amount than Res-FRI-concatenation. We think this comparison also illustrates the advantage of the proposed method when considering differences on computing resources consumption of the two models. In sum, the experiment on CIFAR-10 dataset strongly proves the effectiveness and validity of GoogLe2Net on image classification task and Split-ResFRI also has greatly competitive results when considering the reduction on GFlops and the number of parameters by a significant amount.

\subsection{Experiments on CIFAR-100}
\begin{figure*}[htbp]
	\centering
	\includegraphics[width = 1\textwidth]{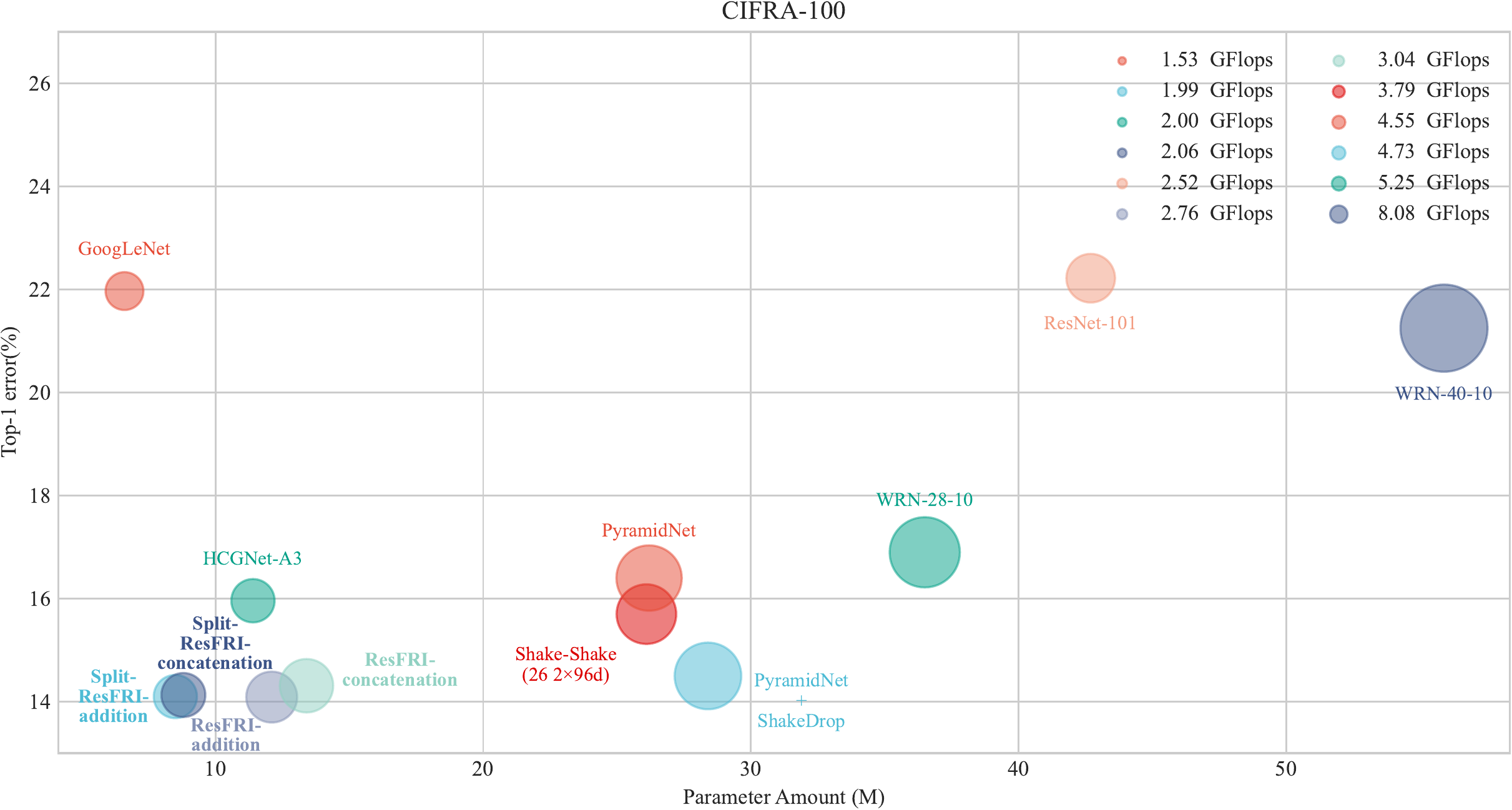}
	\caption{Comparisons of models on CIFAR100 Dataset }
	\label{CIFAR100}
\end{figure*}

The CIFAR100 dataset consists of $50$k training images and $10$k testing images from $100$ classes and their resolution is $32$$\times$$32$. And the detail results of comparisons of different models will be clearly provided in Table \ref{table22} and Fig.\ref{CIFAR100}. 

\begin{table*}[htbp]\footnotesize
	\centering
	\caption{Top-1, Top-5 Test Error (\%) and Model Size on the CIFAR-100 Dataset}
	\renewcommand\arraystretch{1.4} 
	\begin{tabular}{cccccccc} 
		\hline 
		\multicolumn{4}{l}{Model}& Flops &  Params  &    top-1 err. &    top-5 err. \\ \cline{1-8}
		\multicolumn{4}{l}{ResNet-101 \cite{DBLP:conf/cvpr/HeZRS16}}&2.52 GFlops &    42.7M    &      22.22        &5.61\\
		\multicolumn{4}{l}{ResNeXt-50 \cite{DBLP:conf/cvpr/XieGDTH17}}&- &     14.8M    &      22.23        &6.00\\
		\multicolumn{4}{l}{ResNeXt-101 \cite{DBLP:conf/cvpr/XieGDTH17}}&- &     25.3M    &      22.22        &5.99\\
		\multicolumn{4}{l}{ResNeXt-152 \cite{DBLP:conf/cvpr/XieGDTH17}} &  -&   33.3M    &      22.40        &5.58\\
		\multicolumn{4}{l}{DenseNet (k=12, depth=40) \cite{DBLP:conf/cvpr/HuangLMW17}} & - &  1.0M    &      27.55       &-\\
		\multicolumn{4}{l}{DenseNet (k=12, depth=100)\cite{DBLP:conf/cvpr/HuangLMW17}} &   -& 7.0M    &      23.79       &-\\
		\multicolumn{4}{l}{DenseNet (k=24, depth=100)\cite{DBLP:conf/cvpr/HuangLMW17}}&-&    27.2M    &      23.42        &-\\
		\multicolumn{4}{l}{DenseNet-BC (k=24) \cite{DBLP:conf/cvpr/HuangLMW17}}&- &    15.3M     &      19.64        &-\\
		\multicolumn{4}{l}{GoogLeNet \cite{DBLP:conf/cvpr/SzegedyLJSRAEVR15}}&1.53 GFlops &     6.6M    &      21.97        &5.94\\
		\multicolumn{4}{l}{Inception v3 \cite{DBLP:conf/cvpr/SzegedyVISW16}}&- &     22.3M    &      22.81        &6.39\\
		\multicolumn{4}{l}{Inception v4 \cite{DBLP:conf/aaai/SzegedyIVA17}}&- &     41.3M    &      24.14        &6.90\\
		\multicolumn{4}{l}{InceptionResnet v2 \cite{DBLP:conf/aaai/SzegedyIVA17}}&- &     65.4M    &      27.51        &9.11\\
		\multicolumn{4}{l}{Xception \cite{DBLP:conf/cvpr/Chollet17}}&- &     21.0M    &      25.07        &7.32\\
		\multicolumn{4}{l}{WRN-40-10 \cite{DBLP:conf/bmvc/ZagoruykoK16}}&8.08 GFlops &     55.9M    &      21.25        &5.77\\
		\multicolumn{4}{l}{	NIN + Dropout \cite{DBLP:journals/corr/LinCY13}}&- &     0.96M   &      35.68 & -      \\
		\multicolumn{4}{l}{	FitNet-LSUV \cite{DBLP:journals/corr/MishkinM15}}&- &     0.3M   &      29.96 & -      \\
		\multicolumn{4}{l}{	FractalNet \cite{DBLP:conf/iclr/LarssonMS17}}&- &     38.6M   &      29.05 & -      \\
		\multicolumn{4}{l}{SOPCNN \cite{DBLP:conf/mldm/Assiri19}}&- &    4.2M    &      27.04        &-\\
		\multicolumn{4}{l}{WRN-28-10 \cite{DBLP:conf/icml/KwonKPC21}}&5.25 GFlops &     36.5M    &      16.9        &-\\
		\multicolumn{4}{l}{Res2NeXt-29, 6c$\times$24w$\times$6s \cite{DBLP:journals/pami/GaoCZZYT21}}&- &     36.7M    &      16.79        &-\\
		\multicolumn{4}{l}{Res2NeXt-29, 6c$\times$24w$\times$6s-SE \cite{DBLP:journals/pami/GaoCZZYT21}}&- &     36.9M    &      16.56        &-\\
		\multicolumn{4}{l}{PyramidNet \cite{DBLP:conf/cvpr/HanKK17}}&4.55 GFlops &     26.2M    &      16.4       &-\\
		\multicolumn{4}{l}{OR-WideResNet \cite{DBLP:conf/cvpr/ZhouYQJ17}}&- &     18.2M    &      16.15        &2.98\\
		\multicolumn{4}{l}{NASNet-A \cite{DBLP:conf/aaai/YangAZHZXLX20}}&- &     50.9M    &      16.03        &-\\
		\multicolumn{4}{l}{HCGNet-A3 \cite{DBLP:conf/aaai/YangAZHZXLX20}}&2.0 GFlops &     11.4M    &     15.96        &-\\
		\multicolumn{4}{l}{Shake-Shake (26 2$\times$96d) \cite{DBLP:conf/iclr/ForetKMN21}}&3.79 GFlops &     26.1M    &     15.7        &-\\
		\multicolumn{4}{l}{PyramidNet+ShakeDrop \cite{DBLP:conf/iclr/ForetKMN21}}&4.73 GFlops &     28.4M    &     14.5     &-\\\cline{1-8}
		
		\multicolumn{4}{l}{ResFRI-addition}&2.76 GFlops &    12.2M 	    &      \textbf{14.09}        &2.42\\
		\multicolumn{4}{l}{Split-ResFRI-addition}& 1.99 GFlops &      8.5M   &      14.10& 2.48      \\
		\multicolumn{4}{l}{ResFRI-concatenation}&3.04 GFlops &     13.5M     &      14.31 & 2.71 \\
		\multicolumn{4}{l}{Split-ResFRI-concatenation}&2.06 GFlops &     8.8M     &   14.13 & \textbf{2.32}\\
		\hline
	\end{tabular}
	\label{table22}    
\end{table*}

By checking the results given in Table \ref{table22}, some conclusions can be made. ResNet-101 has a performance lag of around $8$\% compared with the proposed method and it utilizes approximate flops and nearly three times parameter amount of ResFRI. For ResNext-series models, all of them achieves analogous performance as ResNet-101 with much less flops and parameter amounts. The situation of DenseNets is also similar, they further reduces the size and computational complexity of the model, but the accuracy of it is still at a comparatively low level. The best accuracy of them has at least a performance disadvantage of more than $5$\% compared with ResFRI-series models. Besides, the inception-series models also have a relatively excellent performance. Particularly, GoogLeNet possesses only $6.6$M parameter amount but achieves an effect that ranks at the top of many models. For NIN, FitNet and SOPCNN, all of the three models can obtain better results on smaller datasets, but they can not acquire desirable results on more convincing datasets like CIFAR-100. Considering the results of WRN-28-10 provided in \cite{DBLP:conf/icml/KwonKPC21}, it achieves a performance leap with a top-1 error rate of about $16$\% and dose not increase flops and parameters amount too much compared with the previous models. And it can be obtained that Res2NeXt can reach a similar performance with roughly the same number of parameters as WRN-28-10. Certainly, ResFRI and Split-ResFRI have higher accuracy with much lower flops and parameter amounts compared with the two categories of models we just discussed. 

Moreover, when considering other modern models, the HCGNet-A3 has a very approxmate flops and parameter amount with GoogLe2Net which realizes nearly two more percent accuracy improvement on classification tasks. For PyramidNet, NASNet-A, Shake-Shake (26 2$\times$96d) and PyramidNet+ShakeDrop, ResFRI and Split-ResFRI still achieve better performances while using less flops and parameter amount. The most light one, Split-ResFRI-addition, can achieve almost the best performance with less than $9$M parameter amount and $2$ Gflops which are between a half and a third of the scales of the four models mentioned before. Especially, PyramidNet+ShakeDrop has the closest effect to the proposed method while possessing $55$\% higher parameter amount and $110$\% more flops than the proposed models at least. Compared with the original PyramidNet, the combination of PyramidNet+ShakeDrop obtains a performance improvement of about $2$\%, which illustrates the possibility of follow-up work using this technology and the effectiveness of ShakeDrop. In sum, based on experimental results provided in Table \ref{table22}, it can be concluded that the proposed method possesses a far better precision on classification task when compared with classical networks. Except for GoogLeNet and DenseNet, all of the other models have larger parameter amount than the proposed model but could not reach a similar level of accuracy, which demonstrates the efficiency and effectiveness of GoogLe2Net. Although GoogLeNet and DenseNet with specific settings own smaller model scale than ResFRI and Split-ResFRI, but our proposed method has a huge advantage in accuracy. Concretely, the version of addition of ResFRI reaches a top-1 error rate $14.09$ and top-5 error rate $2.42$ on CIFAR-100 dataset, in the meantime, Split-ResFRI could achieve very similar performance with at most $37$\% reduction of parameter amount and $34.5$\% curtailment on flops. In one word, all of the comparisons proves the superiority of GoogLe2Net on classification tasks which can be regarded as a satisfying solution in choices among CNN architectures.

\subsection{Experiments on Tiny Imagenet}
The Tiny Imagenet dataset consists of $100$k training images and $10$k testing images from $200$ classes and their resolution is $64$$\times$$64$. And the results of comparisons are given in Table \ref{table2}. And it is worth noting that flops and parameter amounts of ResFRI and Split-ResFRI are evaluated using a tensor matrix of $3\times64\times64$ and the model is subtly adjusted to fit the different type of data, so the number of them will also variate accordingly.

\begin{table}[htbp]
	\centering
	\caption{Top-1 Test Error (\%) and Model Size on the Tiny Imagenet Dataset}
	\renewcommand\arraystretch{1.2} 
	\begin{tabular}{ccccccc} 
		\hline 
		\multicolumn{4}{l}{Model}& Flops &  Params  &    top-1 err.  \\ \cline{1-7}
		\multicolumn{4}{l}{ResNet-18+Mixup+DM \cite{https://doi.org/10.48550/arxiv.2203.10761}}&- &   11.1M   &  34.93   \\
		\multicolumn{4}{l}{ResNet-18+CutMix+DM \cite{https://doi.org/10.48550/arxiv.2203.10761}}&- &   11.1M  &  33.55  \\
		\multicolumn{4}{l}{ResNet-18+ManifoldMix+DM \cite{https://doi.org/10.48550/arxiv.2203.10761}}&- &   11.1M   &  34.55  \\
		\multicolumn{4}{l}{ResNet-18+ResizeMix+DM \cite{https://doi.org/10.48550/arxiv.2203.10761}}&- &   11.1M   &  35.67   \\
		\multicolumn{4}{l}{ResNet-18+PuzzleMix+DM \cite{https://doi.org/10.48550/arxiv.2203.10761}}&- &   11.1M   &  33.48   \\
		
		\multicolumn{4}{l}{ResNeXt-50+Mixup+DM \cite{https://doi.org/10.48550/arxiv.2203.10761}}&- &   23.3M   &  32.30   \\
		\multicolumn{4}{l}{ResNeXt-50+CutMix+DM \cite{https://doi.org/10.48550/arxiv.2203.10761}}&- &   23.3M   &  32.54  \\
		\multicolumn{4}{l}{ResNeXt-50+ManifoldMix+DM \cite{https://doi.org/10.48550/arxiv.2203.10761}}&- &   23.3M   &  31.52   \\
		\multicolumn{4}{l}{ResNeXt-50+ResizeMix+DM \cite{https://doi.org/10.48550/arxiv.2203.10761}}&- &   23.3M   &  31.44   \\
		\multicolumn{4}{l}{ResNeXt-50+PuzzleMix+DM \cite{https://doi.org/10.48550/arxiv.2203.10761}}&- &   23.3M   &  31.96   \\
		\multicolumn{4}{l}{WaveMixLite-144/7 \cite{DBLP:journals/corr/abs-2205-14375}}&- &    3.01 M    &      47.62        \\
		\multicolumn{4}{l}{DenseNet + Residual Networks \cite{DBLP:journals/corr/abs-1904-10429}}&- &     -    &      40.0      \\
		\multicolumn{4}{l}{ResNet18 + AutoMix \cite{DBLP:conf/eccv/ZhuSYZ20}}&- &     11.1M   &   32.67        \\
		\multicolumn{4}{l}{UPANets \cite{DBLP:journals/corr/abs-2103-08640}}&- &     24.4M   &     32.33          \\
		\multicolumn{4}{l}{ResNet18 + SAMix \cite{DBLP:journals/corr/abs-2111-15454}}&- &     11.1M   &   31.11        \\
		\multicolumn{4}{l}{	PreActResNet-18-3 + MixMo \cite{DBLP:conf/iccv/RameSC21}}&- &     11.1M   &     29.76          \\
		\cline{1-7}
		
		\multicolumn{4}{l}{ResFRI-addition (pruning ratio $0.7$)}&3.13 GFlops &    12.4M 	    &    31.5        \\
		\multicolumn{4}{l}{ResFRI-addition (pruning ratio $0$)}&3.13 GFlops &    12.4M 	    &      29.60      \\
		\multicolumn{4}{l}{Split-ResFRI-addition}& 2.37 GFlops&   8.5M   &   31.93    \\
		\multicolumn{4}{l}{ResFRI-concatenation}&3.4 GFlops &     13.7M     &      \textbf{29.46}  \\
		\multicolumn{4}{l}{Split-ResFRI-concatenation}&2.44 GFlops & 9.0M &      32.04 \\
		\hline
	\end{tabular}
	\label{table2}    
\end{table}

The experiments on the Tiny Imagenet show that the proposed method achieves a considerably satisfying classification accuracy. For ResNet-18, it has nearly the same as many parameters as ResFRI, but achieves far weaker performance than ResFRI. Besides, compared with Split-ResFRI, the Split-ResFRI can obtain higher accuracy using less parameter amounts, which clearly demonstrates the efficiency of the proposed model. Moreover, ResNext-50 possesses two to three times as many as parameters as ResFRI and Split-ResFRI, it is able to get approximate performance to the proposed models but still falls behind in the best model accuracy. And with respect to WaveMixLite-144/7, it reaches a similar performance to ResNet-50 utilizing only $3$M parameters. But its actual model accuracy is still not satisfactory. Compared with the methods like ResNet18 and PreActResNet, ResFRIs provide a best performance exceeding $70$\% accuracy which is a remarkable improvement. It is worth noting that Split-ResFRIs are also able to achieve a similar tier of accuracy utilizing less parameter amounts. In sum, GoogLe2Net reaches a high level of performance on image classification task without consuming too many computing resources in comparison with other models.

\subsection{Experiments on MNIST}
The MNIST dataset contains $60$k training images and $10$k testing images from $10$ classes whose resolution is $28$$\times$$28$. And the detail results of comparisons of different models will be clearly provided in Table \ref{table3}. It is worth noting that parameter amounts of ResFRI and Split-ResFRI are calculated using a tensor matrix of $3\times32\times32$, because the images of MNIST are resized into $32\time32$ before being inputting proposed models for process of training.


\begin{table}[htbp]
	\centering
	\caption{Test Accuracy (\%) and Model Size on the MNIST Dataset}
	\renewcommand\arraystretch{1.2} 
	\begin{tabular}{cccccc} 
		\hline 
		\multicolumn{4}{l}{Model} &  Params  &    top-1 err. \\ \cline{1-6}
		\multicolumn{4}{l}{	PCANET-1 \cite{DBLP:journals/tip/ChanJGLZM15}} &     -   &      0.62      \\
		\multicolumn{4}{l}{	FitNet-LSUV \cite{DBLP:journals/corr/MishkinM15}} &     0.3M   &      0.46      \\
		\multicolumn{4}{l}{	NiN \cite{DBLP:journals/corr/LinCY13}} &     0.96M   &      0.45      \\
		\multicolumn{4}{l}{VGG8B \cite{DBLP:conf/icml/NoklandE19}} &     7.3M   &      0.26    \\
		\multicolumn{4}{l}{	CapsNet \cite{DBLP:conf/nips/SabourFH17}} &     -   &      0.25    \\
		\multicolumn{4}{l}{	DropConnect \cite{DBLP:conf/icml/WanZZLF13}} &     -   &      0.21       \\
		\multicolumn{4}{l}{	RMDL \cite{DBLP:conf/icisdm/KowsariHBMB18}} &     -   &      0.18       \\
		\multicolumn{4}{l}{	SOPCNN \cite{DBLP:conf/mldm/Assiri19}} &     1.4M    &      0.17      \\
		\cline{1-6}
		\multicolumn{4}{l}{ResFRI-addition} &     12.1M    &      0.35      \\
		\multicolumn{4}{l}{Split-ResFRI-addition}&       8.4M   &      0.39    \\
		\multicolumn{4}{l}{ResFRI-concatenation} &     13.4M    &     \textbf{0.31}        \\
		\multicolumn{4}{l}{Split-ResFRI-concatenation} &     8.8M &    0.35 \\
		\hline
	\end{tabular}
	\label{table3}    
\end{table}

Based on MNIST dataset, there exist many very light models which still reach a great level of accuracy. The proposed model falls behind by approximately $0.1$ to $0.18$ percent and consumes much more computing resources. Nevertheless, FitNet-LSUV and NiN encounter more than $4$ and $15$ percent performance loss on CIFAR-10 and CIFAR-100 dataset provided in Table \ref{table1} and \ref{table22} respectively compared with GoogLe2Net, which demonstrates that the relatively lower level of precision of GoogLe2Net on MNIST dataset is completely acceptable. Besides, the remaining methods like CapsNet, RMDL and SOPCNN also have similar situations. Thus, the proposed method is more comprehensive and universal in handling classification tasks. And with respect to the performances of the two version of ResFRI, we argue that because the features contained in MNIST are simpler comparatively, the operation of concatenation is helpful to strengthen features instead of constructing too dense connection between convolutional layers. Moreover, for Split-ResFRIs, the performances of them become a little weaker in comparison with the versions without split, which may be caused by reduction of feature extraction operations.

\subsection{Experiments on FashionMNIST}
The FashionMNIST dataset consists of $60$k training images and $10$k testing images from $10$ classes and their resolution is $28$$\times$$28$.  And the results of comparisons are given in Table \ref{table4}. It is worth noting that parameter amounts of ResFRI and Split-ResFRI are calculated using a tensor matrix of $3\times32\times32$, because the images of FashionMNIST are resized into $32\time32$ before being inputting proposed models for process of training.


\begin{table}[htbp]
	\centering
	\caption{Test Accuracy (\%) and Model Size on the FashionMNIST Dataset\\}
	\renewcommand\arraystretch{1.2} 
	\begin{tabular}{cccccc} 
		\hline 
		\multicolumn{4}{l}{Model} &  Params  &    top-1 err. \\ \cline{1-6}
		\multicolumn{4}{l}{Inception v3 \cite{DBLP:journals/corr/abs-2203-15331}} &     24.7M     &      5.56      \\
		\multicolumn{4}{l}{SeResNeXt101-STD \cite{DBLP:conf/cvpr/HuSS18}} &     -     &      4.59       \\
		\multicolumn{4}{l}{VGG8B(2x) \cite{DBLP:conf/icml/NoklandE19}} &     28M     &      4.33      \\
		\multicolumn{4}{l}{PreAct-ResNet18 \cite{DBLP:conf/eccv/HeZRS16}} &     11.1M     &      4.30      \\
		\multicolumn{4}{l}{WideResNet-28-10 \cite{DBLP:conf/icml/NoklandE19}} &     37M     &      4.16       \\
		\multicolumn{4}{l}{DenseNet-BC-190 \cite{DBLP:conf/cvpr/HuangLMW17}} &     25.6M     &      4.06      \\
		\cline{1-6}
		\multicolumn{4}{l}{ResFRI-addition} &     12.1M    &      4.00        \\
		\multicolumn{4}{l}{Split-ResFRI-addition}&       8.4M   &      \textbf{3.80}     \\
		\multicolumn{4}{l}{ResFRI-concatenation} &     13.4M    &     4.29        \\
		\multicolumn{4}{l}{Split-ResFRI-concatenation} &     8.8M     &   3.87 \\
		\hline
	\end{tabular}
	\label{table4}    
\end{table}

As shown in the Table \ref{table4}, ResFRI and Split-ResFRI reach a satisfying level of accuracy on FashionMNIST dataset. And ResFRI-addition and ResFRI-concatenation make $1.56$\% and $1.28$\% percent performance gains compared with the Inception v3, which proves the efficiency and effectiveness of ResFRI compared with other Inception-like architecture. Besides, ResFRI and Split-ResFRI also outstrips these traditional models such as WideResNet, VGG8B and DenseNet utilizing much less parameter amount. And it is worth noting that Split-ResFRI outperforms ResFRI on FashionMNIST dataset, which is very interesting and probably tells us that extraction of picture features like simple objects don't require a deep and dense neural network. All in all, by checking the results of comparison, it can be concluded that the proposed method guarantees a enough precision on a relatively small and simple dataset and splitting features may be helpful in improving performance in analogous tasks.

\subsection{Experiments on SVHN}
The SVHN dataset contains $73257$ training images and $26032$ testing images from $10$ classes whose resolution is $32$$\times$$32$. And the detail results of comparisons of different models will be clearly provided in Table \ref{table5}.


\begin{table}[htbp]
	\centering
	\caption{Test Accuracy (\%) and Model Size on the SVHN Dataset}
	\renewcommand\arraystretch{1.2} 
	\begin{tabular}{cccccc} 
		\hline 
		\multicolumn{4}{l}{Model} &  Params  &    top-1 err. \\ \cline{1-6}
		\multicolumn{4}{l}{FitNet \cite{DBLP:journals/corr/RomeroBKCGB14}} &     -   &      2.42      \\
		\multicolumn{4}{l}{NiN \cite{DBLP:journals/corr/LinCY13}} &     0.96M   &      2.35      \\
		\multicolumn{4}{l}{FractalNet \cite{DBLP:conf/iclr/LarssonMS17}} &     38.6M    &      2.01      \\
		\multicolumn{4}{l}{DropConnect \cite{DBLP:conf/icml/WanZZLF13}} &     -    &      1.94     \\
		\multicolumn{4}{l}{Deeply Supervised Net \cite{DBLP:conf/aistats/LeeXGZT15}} &     -    &      1.92      \\
		\multicolumn{4}{l}{FractalNet with Dropout/Drop-path \cite{DBLP:conf/iclr/LarssonMS17}} &     38.6M    &      1.87      \\
		\multicolumn{4}{l}{ResNet with Stochastic Depth \cite{DBLP:conf/eccv/HuangSLSW16}} &     1.7M    &      1.75      \\
		\multicolumn{4}{l}{DenseNet-BC \cite{DBLP:conf/cvpr/HuangLMW17}} &     15.3M    &      1.74     \\\cline{1-6}
		
		\multicolumn{4}{l}{ResFRI-addition} &     12.1M    &      \textbf{1.72}        \\
		\multicolumn{4}{l}{Split-ResFRI-addition} &      8.4M   &      1.84     \\
		\multicolumn{4}{l}{ResFRI-concatenation} &     13.4M    &      1.75      \\
		\multicolumn{4}{l}{Split-ResFRI-concatenation} &     8.8M     &   1.82 \\
		\hline
	\end{tabular}
	\label{table5}    
\end{table}

By analyzing the experimental results on SVHN dataset, the GoogLe2Net also achieves relatively satisfying accuracy. Compared with the classical models like NiN, FractalNet and DenseNet, the proposed method utilizes much less parameter amount to reach a similar level of precision. Especially, the FractalNet possesses $219$\% higher parameter amount than GoogLe2Net while falling behind by $0.15$ percent accuracy compared with ResFRI. And it's worth noting that the performance of proposed model also exceeds FractalNet and DenseNet on CIFAR-10 and CIFAR-100 dataset.

\begin{table}[htbp]
	\centering
	\caption{Comparison among ResFRI variants on CIFAR10 dataset}
	\renewcommand\arraystretch{1.2} 
	\begin{tabular}{cccccc} 
		\hline 
		 \multicolumn{4}{c}{Variants} &  Params  &    top-1 err. \\ \cline{1-6}
		\multicolumn{4}{c}{\makecell{ResFRI\\ (addition, pruning ratio 0.7)}}&    12.1M    &      \textbf{2.08}       \\
		\multicolumn{4}{c}{\makecell{ResFRI\\ (addition, pruning ratio 0.35)}} &    12.1M    &      2.29      \\
		\multicolumn{4}{c}{\makecell{ResFRI\\ (addition, pruning ratio 0)}}&    12.1M    &      2.13       \\
		\multicolumn{4}{c}{\makecell{ResFRI \\(concatenation, pruning ratio 0.7)}}&     13.4M     &      2.14 \\
		\multicolumn{4}{c}{\makecell{ResFRI \\(concatenation, pruning ratio 0.35)}} &     13.4M     &      2.23 \\
		\multicolumn{4}{c}{\makecell{ResFRI\\ (concatenation, pruning ratio 0)}} &     13.4M     &      \textbf{2.06} \\\cline{1-6}
	    \multicolumn{4}{c}{\makecell{ResFRI without AvgPooling layer\\ (addition, pruning ratio 0.7)}} &     12.1M    &      2.25       \\
	    \multicolumn{4}{c}{\makecell{ResFRI without residual connection\\ (addition, pruning ratio 0.7)}}&     8.9M    &      2.43       \\
	    \multicolumn{4}{c}{\makecell{ResFRI without transverse passages\\ (addition, pruning ratio 0.7)}} &     9.4M   &      2.12       \\
	    \cline{1-6}
	    \multicolumn{4}{c}{\makecell{ResFRI without AvgPooling layer\\ (concatenation, pruning ratio $0$))}} &     13.4M    &      2.30       \\
	    \multicolumn{4}{c}{\makecell{ResFRI without residual connection\\ (concatenation, pruning ratio $0$)}}&     10.2M    &      2.37       \\
	    \multicolumn{4}{c}{\makecell{ResFRI without transverse passages \\(concatenation, pruning ratio $0$)}} &     9.4M   &      2.60       \\	    \hline
        \end{tabular}
     \label{table6}    
\end{table}

\subsection{Ablation Experiment}
In this section, we conduct the ablation experiment from two main aspects which are addition and concatenation version of ResFRI. In the preliminary stage of our experiment, we notice that for the addition version of ResFRI, a proper ration of pruning may help to promote the accuracy of the model. And in the version of concatenation, no pruning may further enhance performance of the network. Therefore, all of the ablation experiments are not only involved with adjustment of structure of networks, but also the ratios of pruning. And all of the results are provided in the following Table \ref{table6}.

In detail, we remove three key components of ResFRI, namely AvgPooling layer, Residual connection and transverse passages between groups of convolutional layers respectively, to verify their influence on performance of the proposed network. And we can notice that when each of them is removed, the performance will encounter a precision loss to some extent. It strongly demonstrates that when all of those components are synthesized, the lowest top1-error can be reached. Moreover, we also test effects of different pruning ratio on precision of the proposed model, which also proves the rationality of our settings of ResFRI on CIFAR-10. 

\section{CONCLUSIONS}
In this paper, we first review the architectures of traditional neural networks and state importance of multi-scale design in CNNs. For the structure of incpetion-like networks, we notice that construction of transverse passages between adjacent groups of convolutional layers may boost performance of the network compared with original inception frameworks. Besides, referring the concept of ResNet, we also adopt a policy that a residual connection is added to lower difficulty in network optimization. In detail, transverse passages between adjacent groups of convolutional layers realize feature reutilization in groups of convolutional layers and further enhance the ability of expression and generalization of original inception. Besides, residual connection reduces overfitting and gradient disappearance. They are the main reasons that GoogLe2Net is able to reach a satisfactory level of accuracy on mainstream vision datasets under such a light and efficient inception-like framework. And all the experiments in this paper confirm this perspective. Moreover, in the future, we believe the organic combination of the concept of multi-scale and CNNs will be a hot spot in boosting performances on various vision tasks.







\bibliographystyle{elsarticle-num}
\bibliography{cite}
\end{document}